# Model Reduction Techniques for Computing Approximately Optimal Solutions for Markov Decision Processes

Thomas Dean and Robert Givan and Sonia Leach
Department of Computer Science, Brown University
{tld, rlg, sml}@cs.brown.edu
http://www.cs.brown.edu/people/

## Abstract

We present a method for solving implicit (factored) Markov decision processes (MDPs) with very large state spaces. We introduce a property of state space partitions which we call $\epsilon$-homogeneity. Intuitively, an $\epsilon$-homogeneous partition groups together states that behave approximately the same under all or some subset of policies. Borrowing from recent work on model minimization in computer-aided software verification, we present an algorithm that takes a factored representation of an MDP and an $0 \leq \epsilon \leq 1$ and computes a factored $\epsilon$-homogeneous partition of the state space.

This partition defines a family of related MDPs—those MDP's with state space equal to the blocks of the partition, and transition probabilities "approximately" like those of any (original MDP) state in the source block. To formally study such families of MDPs, we introduce the new notion of a "bounded parameter MDP" (BMDP), which is a family of (traditional) MDPs defined by specifying upper and lower bounds on the transition probabilities and rewards. We describe algorithms that operate on BMDPs to find policies that are approximately optimal with respect to the original MDP.

In combination, our method for reducing a large implicit MDP to a possibly much smaller BMDP using an $\epsilon$-homogeneous partition, and our methods for selecting actions in BMDP's constitute a new approach for analyzing large implicit MDP's. Among its advantages, this new approach provides insight into existing algorithms to solving implicit MDPs, provides useful connections to work in automata theory and model minimization, and suggests methods, which involve varying $\epsilon$, to trade time and space (specifically in terms of the size of the corresponding state space) for solution quality.

## 1 Introduction

Markov decision processes (MDP) provide a formal basis for representing planning problems involving uncertainty [Boutilier et al., 1995a]. There exist algorithms for solving MDPs that are polynomial in the size of the state space [Puterman, 1994]. In this paper, we are interested in MDPs in which the states are specified implicitly using a set of state variables. These MDPs have explicit state spaces which are exponential in the number of state variables, and are typically not amenable to direct solution using traditional methods due to the size of the explicit state space.

It is possible to represent some MDPs using space polylog in the size of the state space by factoring the state-transition distribution and the reward function into sets of smaller functions. Unfortunately, this efficiency in representation need not translate into an efficient means of computing solutions. In some cases, however, dependency information implicit in the factored representation can be used to speed computation of an optimal policy [Boutilier and Dearden, 1994, Boutilier et al., 1995b, Lin and Dean, 1995].

The resulting computational savings can be explained in terms of finding a *homogeneous* partition of the state space—a partition such that states in the same block transition with the same probability to each of the other blocks. Such a partition induces a smaller, explicit MDP whose states are the blocks of the partition; the smaller MDP, or *reduced model* is equivalent to the original MDP in a well defined sense. It is possible to take an MDP in factored form and find its smallest reduced model using a number of "partition splitting" operations polynomial in the size of the resulting model; however, these splitting operations are in general propositional logic operations which are $\mathcal{NP}$-hard and are thus only heuristically effective. The states of the reduced process correspond to groups of states (in the original process) that behave the same under all policies. The original and reduced processes are equivalent in the sense that they yield the same solutions, i.e., the same optimal policies and state values.

The basic idea of computing equivalent reduced pro-cesses has its origins in automata theory [Hartmanis and Stearns, 1966] and stochastic processes [Kemeny and Snell, 1960] and has surfaced more recently in the work on model checking in computer-aided verification [Burch et al., 1994][Lee and Yannakakis, 1992]. Building on the work of Lee and Yannakakis [1992], we have shown [Dean and Givan, 1997] that several existing algorithms are asymptotically equivalent to first constructing the minimal reduced MDP and then solving this MDP using traditional methods that operate on the flat (unfactored) representations.

The minimal model may be exponentially larger than the original compact MDP. In response to this problem, this paper introduces the concept of an $\epsilon$-homogeneous partition of the state space. This relaxation of the concept of homogeneous partition allows states within the same block to transition with different probabilities to other blocks so long as the different probabilities are within $\epsilon$. For $\epsilon > 0$, there are generally $\epsilon$-homogeneous partitions which are smaller and often much smaller than the smallest homogeneous partition. In this paper we discuss *approximate model reduction*—an algorithm for finding an $\epsilon$-homogeneous partition of a factored MDP which is generally smaller and always no larger than the smallest homogeneous partition.

Any $\epsilon$-homogeneous partition induces a family of explicit MDPs, each with state space equal to the blocks of the partition, and transition probabilities from each block nearly identical to those of the underlying states. To formalize and analyze such families we introduce the new concept of a *bounded parameter MDP* (BMDP)—an MDP in which the transition probabilites and rewards are given not as point values but as closed intervals. In Givan et al. [1997], we describe algorithms that operate on BMDPs to produce bounds on value functions and thereby compute approximately optimal policies—we summarize these methods here. The resulting bounds and policies apply to the original implicit MDP. Bounded parameter MDPs generalize traditional (exact) MDPs and are related to constructs found in work on aggregation methods for solving MDPs [Schweitzer, 1984, Schweitzer et al., 1985, Bertsekas and Castañon, 1989]. Although BMDPs are introduced here to represent approximate aggregations, they are interesting in their own right and are discussed in more detail in [Givan et al., 1997], The model reduction algorithms and bounded parameter MDP solution methods can be combined to find approximately optimal solutions to large factored MDPs, varying $\epsilon$ to trade time and space for solution quality.

The remainder of this paper is organized as follows. In Section 2, we give an overview of the algorithms and representations in this paper and discuss how they fit together. Section 3 reviews traditional and factored MDPs and describes the generalization to bounded parameter MDPs. Section 4 describes an algorithm for $\epsilon$-reducing an MDP to a (possibly) smaller explicit BMDP (an MDP if $\epsilon = 0$). Section 5 summarizes our methods for policy selection in BMDPs, and addresses the applicability of the selected policies to any MDP which $\epsilon$-reduces to the analyzed BMDP. The remaining sections summarize preliminary experimental results and discuss related work.

## 2 Overview

Here we survey and relate the basic mathematical objects and operations defined later in this paper. We start with a Markov decision process (MDP) $M$ for which we would like to compute an optimal or near optimal policy. Figure 1.a depicts the MDP $M$ as a directed graph corresponding to the state-transition diagram, and its optimal policy $\pi_M^*$ as found by traditional value iteration.

We assume that the state space for $M$ (and hence the state-transition graph) is quite large. We therefore assume that the states of $M$ are encoded in terms of state variables which represent aspects of the state; an assignment of values to all of the state variables constitutes a complete description of a state. In this paper, we assume that the factored representation is in the form of a Bayesian network, such as that depicted in Figure 1.b with four state variables $\{A, B, C, D\}$.

We speak about operations involving $M$, but in practice all operations will be performed symbolically using the factored representation: we manipulate sets of states represented as formulas involving the state variables.

Figure 1.c and Figure 1.d depict the unique smallest homogeneous partition of the state space of $M$, where the blocks are represented (respectively) implicitly and explicitly. The process of finding this partition is called (exact) model minimization. Factored model minimization involves manipulating boolean formulas and is $\mathcal{NP}$-hard, but heuristic manipulation may rarely achieve this worst case.

The smallest homogeneous partition may be exponentially large, so we seek further reduction (at a cost of only approximately optimal solutions) by finding a smaller $\epsilon$-homogeneous partition, depicted in Figure 1.e and Figure 1.f where the blocks are again represented (respectively) implicitly and explicitly.

Any $\epsilon$-homogeneous partition can be used to create a bounded parameter MDP, shown in Figure 1.g and notated as $\mathcal{M}$ —to do this, we treat the partition blocks as (aggregate) states and summarize everything that we know about transitions between blocks in terms of closed real intervals that describe the variation within a block of the transition probabilities to other blocks, i.e., for any action and pair of blocks, we record the upper and lower bounds on the probability of starting in a state in one block and ending up in the other block.[1]

---

[1]The BMDP $\mathcal{M}$ naturally represents a family of MDPs,

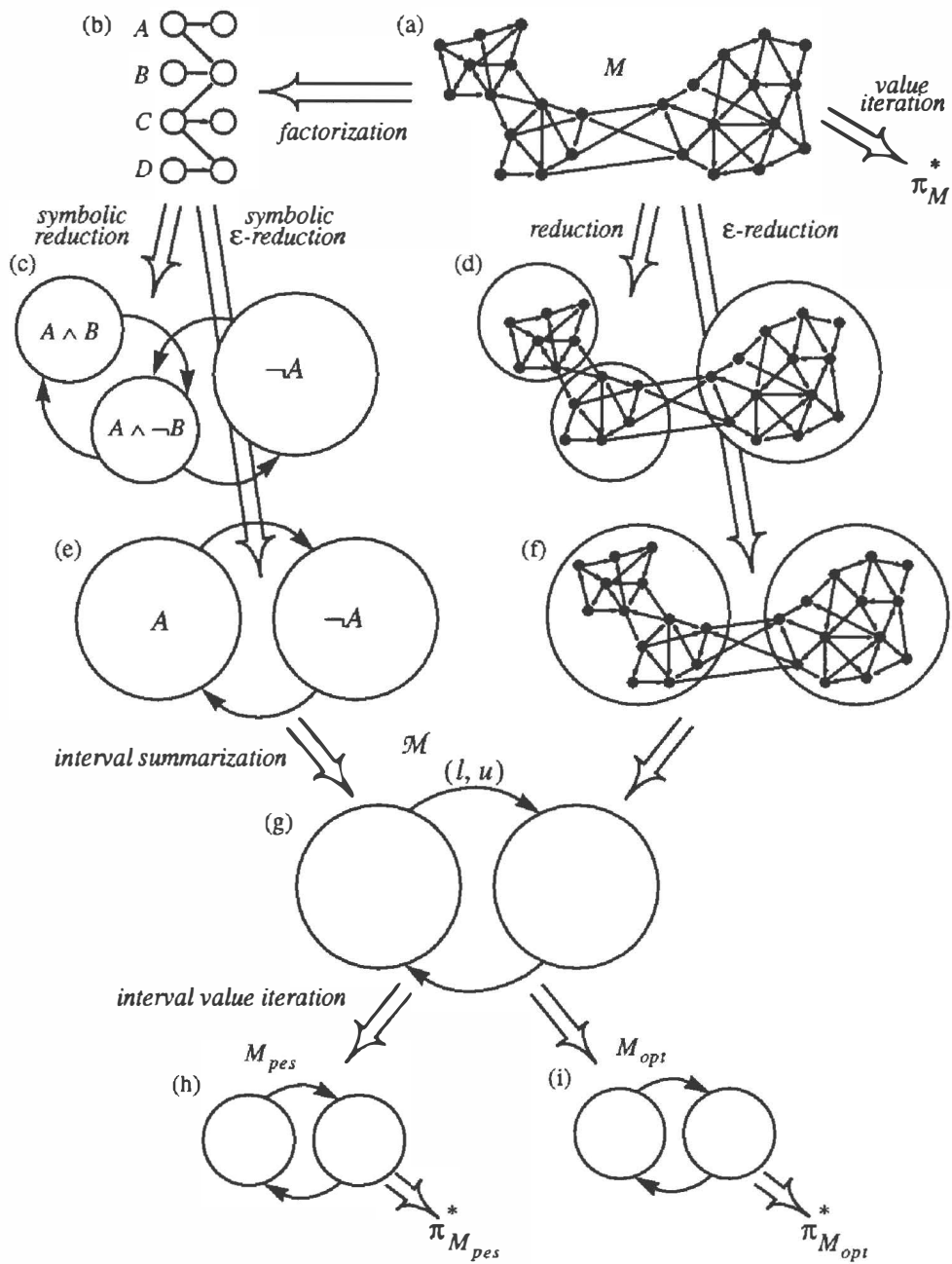

Figure 1: The basic objects and operations described in this paper: (a) depicts the state-transition diagram for an MDP $M$ (only a single action is shown), (b) depicts a Bayesian network as an example of a symbolic representation compactly encoding $M$, (c) and (d) depict the smallest homogeneous partition in (respectively) its implicit (symbolic) and explicit forms, similarly, (e) and (f) depict an $\epsilon$-homogeneous partition in its implicit and explicit forms, (g) represents the bounded-parameter MDP $\mathcal{M}$ summarizing the variations in the $\epsilon$-homogeneous partition, and, finally, (h), (i), and (j) depict particular (exact) MDPs from the family of MDPs defined by $\mathcal{M}$.

Our BMDP analysis algorithms extract particular MDPs from $\mathcal{M}$ that have intuitive characterizations. The *pessimistic model* $M_{pes}$ is the MDP within $\mathcal{M}$ which yields the lowest optimal value $V^*_{M_{pes}}$ at every state. It is a theorem that $M_{pes}$ is well-defined, and that $V^*_{M_{pes}}$ at each state in $\mathcal{M}$ is a lower bound for following the optimal policy $\pi^*_{M_{pes}}$ in any MDP in $\mathcal{M}$ (as well as in the original $M$ from any state in the corresponding block). Similarly, the *optimistic model* $M_{opt}$ has the best value function $V_{M_{opt}}$. $V_{M_{opt}}$ gives upper-bounds for following any policy in $M$. In summary, $V^*_{M_{pes}}$ and $V^*_{M_{opt}}$ give us lower and upper bounds on the optimal value function we are really interested in, $V^*_M$, and following $\pi^*_{M_{pes}}$ in $M$ is guaranteed to achieve at least the lower bound.

Now, armed with this high-level overview to serve as a road map, we descend into the details.

## 3  Markov Decision Processes

**Exact Markov Decision Processes** An (exact) Markov decision process $M$ is a four tuple $M = (\mathcal{Q}, \mathcal{A}, F, R)$ where $\mathcal{Q}$ is a set of states, $\mathcal{A}$ is a set of actions, $R$ is a reward function that maps each state to a real value $R(q)$,[2] $F$ assigns a probability to each state transition for each action, so that for $\alpha \in \mathcal{A}$ and $p, q \in \mathcal{Q}$,

$$F_{pq}(\alpha) = \Pr(X_{t+1} = q | X_t = p, U_t = \alpha)$$

where $X_t$ and $U_t$ are random variables denoting, respectively, the state and action at time $t$.

A *policy* is a mapping from states to actions, $\pi : \mathcal{Q} \to \mathcal{A}$. The *value function* $V_{\pi,M}$ for a given policy maps states to their expected discounted cumulative reward given that you start in that state and act according to the given policy:

$$V_{\pi,M}(p) = R(p) + \gamma \sum_{q \in \mathcal{Q}} f_{pq}(\pi(p)) V_{\pi,M}(q)$$

where $\gamma$ is the *discount rate*, $0 \leq \gamma < 1$. [Puterman, 1994].

**Bounded Parameter MDPs** A *bounded parameter MDP* (BMDP) is a four tuple $\mathcal{M} = (\mathcal{Q}, \mathcal{A}, \hat{F}, \hat{R})$ where $\mathcal{Q}$ and $\mathcal{A}$ are as for MDPs, and $\hat{F}$ and $\hat{R}$ are analogous to the MDP $F$ and $R$ but yield closed real intervals instead of real values. That is, for any action $\alpha$ and states $p, q$, $\hat{R}(p)$ and $\hat{F}_{p,q}(\alpha)$ are both closed real intervals of the form $[l, u]$ for $l$ and $u$ real numbers with $0 \leq l \leq u \leq 1$. For convenience, we define $\underline{F}$ and $\overline{F}$ to be real valued functions which give the lower and upper bounds of the intervals; likewise for $\underline{R}$ and $\overline{R}$.[3] To ensure that $\hat{F}$ admits well-formed transition functions, we require that, for any action $\alpha$ and state $p$, $\sum_{q \in \mathcal{Q}} \underline{F}_{p,q}(\alpha) \leq 1 \leq \sum_{q \in \mathcal{Q}} \overline{F}_{p,q}(\alpha)$.

A BMDP $\mathcal{M} = (\mathcal{Q}, \mathcal{A}, \hat{F}, \hat{R})$ defines a set of exact MDPs $\mathcal{F}_\mathcal{M} = \{M | \mathcal{M} \models M\}$ where $\mathcal{M} \models M$ iff $M = (\mathcal{Q}, \mathcal{A}, F, R)$ and $F$ and $R$ satisfy the bounds provided by $\hat{F}$ and $\hat{R}$ respectively. We will write of *bounding the (optimal or policy specific) value* of a state in a BMDP—by this we mean providing an upper or lower bound on the corresponding state value over the entire family of MDPs $\mathcal{F}_\mathcal{M}$. For a more thorough treatment of BMDPs, please see [Givan et al., 1997].

**Factored Representations** In the remainder of this paper, we make use of *Bayesian networks* [Pearl, 1988] to encode implicit (or *factored*) representations; however, our methods apply to other factored representations such as probabilistic STRIPS operators [Kushmerick et al., 1995]. Let $\mathcal{X} = \{X_1, \ldots, X_m\}$ be a set of state variables. We assume the variables are boolean, and refer to them also as *fluents*. We represent the state at time $t$ as a vector $X_t = \langle X_{1,t}, \ldots, X_{m,t} \rangle$ where $X_{i,t}$ denotes the value of the $i$th state variable at time $t$.

The state transition probabilities can be represented using Bayes networks. A *two-stage temporal Bayesian network* (2TBN) is a directed acyclic graph consisting of two sets of variables $\{X_{i,t}\}$ and $\{X_{i,t+1}\}$ in which directed arcs indicating dependence are allowed from the variables in the first set to variables in the second set and between variables in the second set.[Dean and Kanazawa, 1989] The state-transition probabilities are now factored as

$$\Pr(X_{t+1} | X_t, U_t) = \prod_{i=1}^{m} \Pr(X_{i,t+1} | \text{Parents}(X_{i,t+1}), U_t)$$

where $\text{Parents}(X)$ denotes the parents of $X$ in the 2TBN and each of the conditional probability distributions $\Pr(X_{i,t+1} | \text{Parents}(X_{i,t+1}), U_t)$ can be represented as a conditional probability table or as a decision tree—we choose the latter in this paper following [Boutilier et al., 1995b]. We enhance the 2TBN representation to include actions and reward functions; the resulting graph is called an *influence diagram* [Howard and Matheson, 1984].

Figure 2 illustrates a factored representation with three state variables, $\mathcal{X} = \{P, Q, S\}$, and describes the transition probabilities and rewards for a particular action. The factored form of the transition probabilities

---

but note that the original $M$ is not generally in this family. Nevertheless, our BMDP algorithms compute policies and value bounds which can be soundly applied to the original $M$.

[2]The techniques and results in this paper easily generalize to more general reward functions. We adopt a less general formulation to simplify the presentation.

[3]To simplify the remainder of the paper, we assume that the reward bounds are always tight, i.e., that $\underline{R} = \overline{R}$. The generalization to nontrivial bounds on rewards is straightforward.

128  Dean, Givan, and Leach

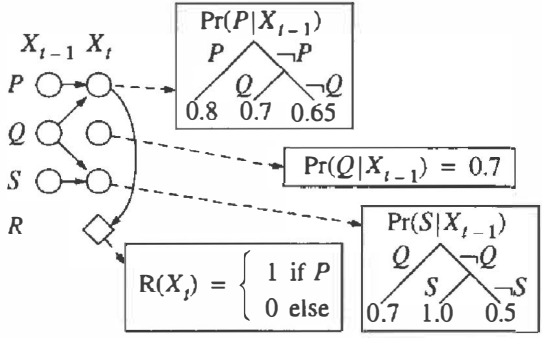

Figure 2: A factored representation with three state variables, $P$, $Q$ and $S$, and reward function $R$.

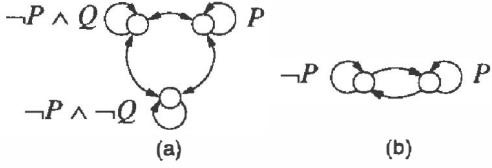

Figure 3: Two $\epsilon$-homogeneous partitions for the MDP described in Figure 2: (a) the smallest exact homogeneous partition ($\epsilon = 0$) and (b) a smaller partition for $\epsilon = 0.05$.

is

$$\Pr(X_{t+1}|X_t, U_t) = \Pr(P_{t+1}|P_t, Q_t) \cdot \Pr(Q_{t+1}) \cdot \Pr(S_{t+1}|S_t, Q_t)$$

where in this case $X_t = \langle P_t, Q_t, S_t \rangle$.

## 4 Model Reduction Methods

In this section, we describe a family of algorithms that take as input an MDP and a real value $\epsilon$ between 0 and 1 and compute a bounded parameter MDP where each closed real interval has extent less than or equal to $\epsilon$. The states in this BMDP correspond to the blocks of a partition of the state space in which states in the same block behave *approximately* the same with respect to the other blocks. The upper and lower bounds in the BMDP correspond to bounds on the transition probabilities (to other blocks) for states that are grouped together.

We first define the property sought in the desired state space partition. Let $\mathcal{P} = \{B_1, \ldots, B_n\}$ be a partition of $\mathcal{Q}$.

**Definition 1** *A partition $\mathcal{P} = \{B_1, \ldots, B_n\}$ of the state space of an MDP $M$ has the property of $\epsilon$-approximate stochastic bisimulation homogeneity with respect to $M$ for $\epsilon$ such that $0 \leq \epsilon \leq 1$ if and only if for each $B_i, B_j \in \mathcal{P}$, for each $\alpha \in \mathcal{A}$, for each $p, q \in B_i$,*

$$|R(p) - R(q)| \leq \epsilon, \quad \text{and}$$

$$\left|\sum_{r \in B_j} F_{pr}(\alpha) - \sum_{r \in B_j} F_{qr}(\alpha)\right| \leq \epsilon$$

*For conciseness, we say $\mathcal{P}$ is $\epsilon$-homogeneous.*[4]

Figure 3 shows two $\epsilon$-homogeneous partitions for the MDP described in Figure 2.

We now explain how we construct an $\epsilon$-homogeneous partition. We first describe the relationship between every $\epsilon$-homogeneous partition and a particular simple partition based on immediate reward.

**Definition 2** *A partition $\mathcal{P}'$ is a refinement of a partition $\mathcal{P}$ if and only if each block of $\mathcal{P}'$ is a subset of some block of $\mathcal{P}$; in this case, we say that $\mathcal{P}$ is coarser than $\mathcal{P}'$, and is a clustering of $\mathcal{P}'$*

**Definition 3** *The immediate reward partition is the partition in which two states, $p$ and $q$, are in the same block if and only if they have the same reward.*

**Definition 4** *A partition $\mathcal{P}$ is $\epsilon$-uniform with respect to a function $f : \mathcal{Q} \to \mathcal{R}$ if for every two states $p$ and $q$ in the same block of $\mathcal{P}$, $|f(p) - f(q)| \leq \epsilon$.*

Every $\epsilon$-homogeneous partition is a refinement of some $\epsilon$-uniform clustering (with respect to reward) of the immediate reward partition. Our algorithm starts by constructing an $\epsilon$-uniform reward clustering $\mathcal{P}_0$ of the immediate reward partition.[5] We then refine this initial partition by splitting[6] blocks repeatedly to achieve $\epsilon$-homogeneity. We can decide which blocks are candidates for splitting using the following local property of the blocks of an $\epsilon$-homogenous partition:

**Definition 5** *We say that a block $C$ of a partition $\mathcal{P}$ is $\epsilon$-stable with respect to a block $B$ iff for all actions $\alpha$ and all states $p \in C$ and $q \in C$ we have*

$$\left|\sum_{r \in B} F_{pr}(\alpha) - \sum_{r \in B} F_{qr}(\alpha)\right| \leq \epsilon$$

*We say that $C$ is $\epsilon$-stable if $C$ is $\epsilon$-stable with respect to every block of $\mathcal{P}$ and action in $\mathcal{A}$.*

The definitions immediately imply that a partition is $\epsilon$-homogenous iff every block in the partition is $\epsilon$-stable.

The *model $\epsilon$-reduction algorithm* simply checks each block for $\epsilon$-stability, splitting unstable blocks until quiescence, *i.e.,* until there are no unstable blocks left to split. Specifically, when a block $C$ is found to be unstable with respect to a block $B$, we replace $C$ in the partition by a set[7] of sub-blocks $C_1, \ldots, C_k$ such that each $C_i$ is a maximal sub-block of $C$ that is $\epsilon$-stable with respect to $B$. *Note that at all times the blocks of the partition are represented in factored form, e.g., as DNF formulas over the state variables. The block splitting operation manipulates these factored representations, not explicit states.* This method is an extension to Markov decision processes of the deterministic model reduction algorithm of Lee and Yannakakis [1992].

---

[4] For the case of $\epsilon = 0$, $\epsilon$-approximate stochastic bisimulation homogeneity is closely related to the *substitution property* for finite automata developed by Hartmanis and Stearns [1966] and the notion of *lumpability* for Markov chains [Kemeny and Snell, 1960].

[5] There may be many such clusterings, we currently choose a coarsest one arbitrarily.

[6] The term *splitting* refers to the process whereby a block of a partition is divided into two or more sub-blocks to obtain a refinement of the original partition.

[7] There may be more than one choice, as discussed below.

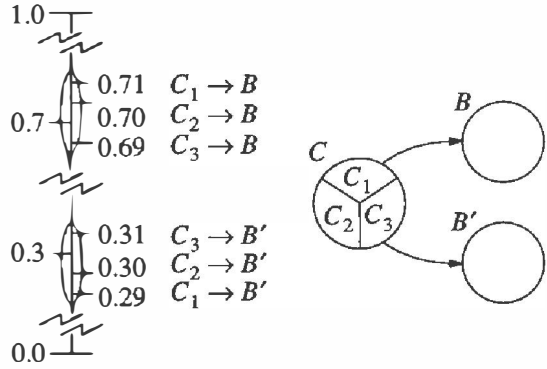

Figure 4: Clustering sub blocks that behave approximately the same. With $\epsilon = 0.01$ there are two smallest clusterings.

If $\epsilon = 0$, the above description fully defines the block splitting operation, as there exists a unique set of maximal, stable sub-blocks. Furthermore, in this case, the algorithm finds the unique smallest homogeneous partition, independent of the order in which unstable blocks are split. We call this partition the *minimal model* (we also use this term to refer to the MDP derived from this partition by treating its blocks as states).

However, if $\epsilon > 0$, then we may have to choose among several possible ways of splitting $C$ as shown in the following example. Figure 4 depicts a block, $C$, and two other blocks, $B$ and $B'$, such that states in $C$ transition to states in $B$ and $B'$ under some action $\alpha$. We partition $C$ into three sub blocks $\{C_1, C_2, C_3\}$ such that states in each sub block have the same transition probabilities with respect to $\alpha$, $B$, and $B'$. In building an 0.01-approximate model, we might replace $C$ by the two blocks $C_1$ and $C_2 \cup C_3$, or by the two blocks $C_3$ and $C_1 \cup C_2$; it is possible to construct examples in which each of these is the most appropriate choice because the splits of other blocks induced later[8]. We require only that the clustering selected is not the refinement of another $\epsilon$-uniform clustering, *i.e.*, that it is as coarse as possible.

Because we make the clustering decisions arbitrarily, our algorithm does not guarantee finding the smallest $\epsilon$-homogenous partition when $\epsilon > 0$, nor that the partition found for $\epsilon_1$ will be smaller (or even as small) as the partition found for $\epsilon_2 < \epsilon_1$. However, it is a theorem that the partition found will be no larger than the unique smallest 0-homogenous partition.

---

[8]The result is additionally sensitive to the order in which unstable blocks are split—splitting one $\epsilon$-unstable block may make another become $\epsilon$-stable.

**Theorem 1** *For $\epsilon > 0$, the partition found by model $\epsilon$-reduction using any clustering technique is coarser than, and thus no larger than the minimal model.*

**Theorem 2** *For $0 < \epsilon_2 < \epsilon_1$, the smallest $\epsilon_1$-homogenous partition is no larger than the smallest $\epsilon_2$-homogenous partition. The model $\epsilon$-reduction algorithm, augmented by an (impractical) search over all clustering decisions, will find these smallest partitions.*

**Theorem 3** *Given a bound and an MDP whose smallest $\epsilon$-homogenous partition is polynomial in size, the problem of determining whether there exists an $\epsilon$-homogenous partition of size no more than the bound is NP-complete.*

These theorems imply that using an $\epsilon > 0$ can only help us, but that our methods may be sensitive to just which $\epsilon$ we choose, and are necessarily heuristic.

Currently our implementation uses a greedy clustering algorithm; in the future we hope to incorporate more sophisticated techniques from the learning and pattern recognition literature to find a smaller clustering locally within each SPLIT operation (though this does not *guarantee* a smaller final partition).

Each $\epsilon$-homogenous partition $\mathcal{P}$ of an MDP $M = (\mathcal{Q}, \mathcal{A}, F, R)$ induces a corresponding BMDP $\mathcal{M}_\mathcal{P} = (\mathcal{Q}, \mathcal{A}, \hat{F}, \hat{R})$ in a straightforward manner. The states of $\mathcal{M}_\mathcal{P}$ are just the blocks of $\mathcal{P}$ and the actions are the same as those in $M$. The reward and transition functions are defined to give intervals bounding the possible reward and block transition probabilities within each block: for blocks $B$ and $C$ and action $\alpha$,

$$\hat{R}(B) = [\ \min_{p \in B} R(p),\ \max_{p \in B} R(p)\ ]$$
$$\hat{F}_{B,C}(\alpha) = [\ \min_{p \in B} \sum_{q \in C} F_{p,q}(\alpha),$$
$$\max_{p \in B} \sum_{q \in C} F_{p,q}(\alpha)\ ]$$

We can then use the methods in the next section to give intervals bounding the optimal value of each state in $\mathcal{M}_\mathcal{P}$ and select a policy which guarantees achieving at least the lower bound value at each state. The following theorem then implies the value bounds apply to the states in $M$, and are achieved or exceeded by following the corresponding policy in $M$.

We first note that any function on the blocks of $\mathcal{P}$ can be extended to a function on the states of $M$: for each state we return the value assigned to the block of $\mathcal{P}$ in which it falls. In this manner, we can interpret the value bounds and policies for $\mathcal{M}_\mathcal{P}$ as bounds and policies for $M$.

**Theorem 4** *For any MDP $M$ and $\epsilon$-homogenous partition $\mathcal{P}$ of the states of $M$, sound (optimal or policy*

specific) value bounds for $\mathcal{M}_\mathcal{P}$ apply also to $M$ (by extending the policy and value functions to the state space of $M$ according to $\mathcal{P}$).

## 5  Interval Value Iteration

We have developed a variant of the value iteration algorithm for computing the optimal policy for exact MDPs[Bellman, 1957] that operates on bounded parameter MDPs. A BMDP $\mathcal{M}$ represents a family of MDPs $\mathcal{F}_\mathcal{M}$, implying some degree of uncertainty as to which MDP in the family actions will actually be taken in. As such, there is no specific value for following a policy from a start state—rather, there is a window of possible values for following the policy in the different MDPs of the family. Similarly, for each state there is a window of possible optimal values over the MDPs in the family $\mathcal{F}_\mathcal{M}$. Our algorithm can compute bounds on policy specific value functions as well as bounds on the optimal value function. We have also shown how to extract from these bounds a specific "optimal" policy which is guaranteed to achieve at least the lower bound value in any actual MDP from the family $\mathcal{F}_\mathcal{M}$ defined by the BMDP. We call this policy $\pi_{\text{pes}}$, the *pessimistic optimal policy*.

We call this algorithm, *interval value iteration (IVI* for optimal values, and $IVI_\pi$ for policy specific values). The algorithm is based on the fact that, if we only knew the rank ordering of the states' values, we would easily be able to select an MDP from the family $\mathcal{F}_\mathcal{M}$ which minimized or maximized those values, and then compute the values using that MDP. Since we don't know the rank ordering of states' values, the algorithm uses the ordering of the current estimates of the values to select a minimizing (maximizing) MDP from the family, and performs one iteration of standard value iteration on that MDP to get new value estimates. These new estimates can then be used to select a new minimizing (maximizing) MDP for the next iteration, and so forth.

Bounded parameter MDPs are interesting objects and we explore them at greater length in [Givan et al., 1997]. In that paper, we prove the following results about $IVI$.

**Theorem 5** *Given a BMDP $\mathcal{M}$ and a specific policy $\pi$, $IVI_\pi$ converges at each state to lower and upper bounds on the value of $\pi$ at that state over all the MDPs in $\mathcal{F}_\mathcal{M}$.*

**Theorem 6** *Given a BMDP $\mathcal{M}$, $IVI$ converges at each state to lower and upper bounds on the optimal value of that state over all the MDPs in $\mathcal{F}_\mathcal{M}$.*

**Theorem 7** *Given a BMDP $\mathcal{M}$, the policy $\pi_{pes}$ extracted by assuming that states actual values are the $IVI$-converged lower bounds has a policy specific lower bound (from $IVI_\pi$) in $\mathcal{M}$ equal to the (non policy specific) $IVI$-converged lower bound. No other policy has a higher policy specific lower bound.*

## 6  Related Work and Discussion

This paper combines a number of techniques to address the problem of solving (factored) MDPs with very large states spaces. The definition of $\epsilon$-homogeneity and the model reduction algorithms for finding $\epsilon$-homogeneous partitions are new, but draw on techniques from automata theory and symbolic model checking. Burch *et al.* [1994] is the standard reference on symbolic model checking for computer-aided design. Our reduction algorithm and its analysis were motivated by the work of Lee and Yannakakis [1992] and Bouajjani *et al.* [1992].

The notion of bounded-parameter MDP is also new, but is related to aggregation techniques used to speed convergence in iterative algorithms for solving exact MDPs. Bertsekas and Castañon [1989] use the notion of aggregated Markov chains and consider grouping together states with approximately the same residuals (*i.e.*, difference in the estimated value function from one iteration to the next during value iteration).

The methods for manipulating factored representations of MDPs were largely borrowed from Boutilier *et al.* [1995b], which provides an iterative algorithm for finding optimal solutions to factored MDPs. Dean and Givan [1997] describe a model-minimization algorithm for solving factored MDPs which is asymptotically equivalent to the algorithm in [Boutilier *et al.*, 1995b].

Boutilier and Dearden [?] extend the work in [Boutilier *et al.*, 1995b] to compute approximate solutions to factored MDPs by associating upper and lower bounds with symbolically represented blocks of states. States are aggregated if they have approximately the same value rather than if they behave approximately the same behavior under all or some set of policies, though it often turns out that states with nearly the same value have nearly the same dynamics.

There are two significant differences between our approximation techniques and those of Boutilier and Dearden. First, we partition the state space and then perform interval value iteration on the resulting bounded-parameter MDP, while Boutilier and Dearden repeatedly partition the state space. Second, we use a fixed $\epsilon$ for computing a partition while Boutilier and Dearden, like Bertsekas and Castañon, repartition the state space (if necessary) on each iteration on the basis of the current residuals, and, hence, (effectively) they use different $\epsilon$'s at different times and on different portions of the state space. Despite these differences, we conjecture that the two algorithms perform asymptotically the same. Practically speaking, we expect that in some cases, repeatedly and adaptively computing partitions may provide better performance, while in other cases, performing the partition once and for all may result in a computational advantage.

We have written a prototype implementation of the model reduction algorithms described in this paper, along with the BMDP evaluation algorithms (IVI) referred to. Using this implementation we have been able to demonstrate substantial reductions in model size, and increasing reductions with increasing $\epsilon$. However, the MDPs we have been reducing are still "toy" problems and while they were not concocted expressly to make the algorithm look good, these empirical results are still of questionable value. Further research is necessary before these techniques are adequate to handle a real-world large scale planning problem in order to give convincing empirical data.

Finally, we believe that by formalizing the notions of approximately similar behavior, approximately equivalent models, and families of closely related MDPs the mathematical entities corresponding to $\epsilon$-homogeneous partitions, $\epsilon$-reductions, and bounded-parameter MDPs provide valuable insight into factored MDPs and the prospects for solving them efficiently.